\documentclass[10pt,twocolumn,letterpaper]{article}

\usepackage{cvpr} 

\usepackage{makecell}
\usepackage{graphics}
\usepackage{epsfig}
\usepackage{newtxtext}
\usepackage{times}
\usepackage{amsmath}
\usepackage{amssymb}
\usepackage{forest}
\usepackage{tikz}
\usepackage{multirow}
\usepackage{graphicx}
\usepackage{array}
\usepackage{tabularx}
\usepackage{caption}
\usepackage{xcolor}
\usetikzlibrary{shapes.geometric}
\usepackage{pifont}
\usepackage{float}

\newcommand{\datasetname}{\textsc{BOP-Ask}}

\newcommand{\cmark}{\textcolor{darkgreen}{\ding{51}}}
\newcommand{\xmark}{\textcolor{darkred}{\ding{55}}}
\definecolor{darkgreen}{RGB}{0,127,0}
\definecolor{darkred}{RGB}{200,0,0}

\usepackage{stfloats}

\definecolor{cvprblue}{rgb}{0.21,0.49,0.74}
\usepackage[pagebackref,breaklinks,colorlinks,allcolors=cvprblue]{hyperref}

\usepackage{graphicx}
\usepackage{array}
\usepackage[margin=1in]{geometry}
\usepackage{ragged2e} 
\usepackage{xcolor}
\definecolor{magline}{HTML}{FF00FF}
\definecolor{blueline}{HTML}{00BFFF}
\definecolor{greenline}{HTML}{01FE03}
\newcommand{\qimg}[2]{%
    \begin{tabular}{@{}m{0.34\linewidth}@{}m{0.66\linewidth}@{}}\footnotesize\RaggedRight #1 & \includegraphics[width=\linewidth]{#2}\end{tabular}}

\graphicspath{{tables/john/}}

\renewcommand{\arraystretch}{1.3}

\def\paperID{31924} 
\def\confName{CVPR}
\def\confYear{2026}

\title{\datasetname: Object-Interaction Reasoning for Vision-Language Models}
\author{
Vineet Bhat$^{1}$ \quad
Sungsu Kim$^{1}$ \quad
Valts Blukis$^{2}$ \quad
Greg Heinrich$^{2}$ \quad
Prashanth Krishnamurthy$^{1}$ \quad\\
Ramesh Karri$^{1}$ \quad
Stan Birchfield$^{2}$ \quad
Farshad Khorrami$^{1}$ \quad
Jonathan Tremblay$^{2}$ \\
\\[-2pt]
$^{1}$New York University \qquad
$^{2}$NVIDIA
}

\begin{document}
\maketitle
\begin{abstract}

%
Vision–Language Models (VLMs) have achieved impressive performance on spatial reasoning benchmarks, yet these evaluations mask critical weaknesses in understanding object interactions. 
Current benchmarks test high-level relationships (``left of," ``behind", etc.) but ignore fine-grained spatial understanding needed for real-world applications: precise 3D localization, physical compatibility between objects, object affordances and multi-step spatial planning.
In this work, we present \datasetname{}, a novel large-scale dataset for object-interaction reasoning for both training and benchmarking.
Our data generation pipeline leverages 6D object poses from the Benchmark for Object Pose Estimation (BOP) datasets
from which we derive fine-grained annotations such as grasp poses, referred object poses, path planning trajectories, relative spatial and depth relationships, and object-to-object relationships.
\datasetname{} comprises over 150k images and 33M question–answer pairs spanning six tasks (four novel), providing a rich resource for training and evaluating VLMs.
We evaluate proprietary and open-sourced VLMs, and conduct human evaluations on BOP-ASK-core, a contributed test benchmark. We also release BOP-ASK-lab, an out-of-distribution benchmark with images not sourced from BOP, enabling testing of generalization. Our experiments demonstrate that models trained on \datasetname{} outperform baselines and exhibit emergent capabilities such as precise object and grasp pose estimation, trajectory planning, and fine-grained object-centric spatial reasoning in cluttered environments.
Project website: \url{https://bop-ask.github.io/} 

\end{abstract}

\section{Introduction}



\begin{figure}[t]
    \centering
    \includegraphics[width=1.05\linewidth]{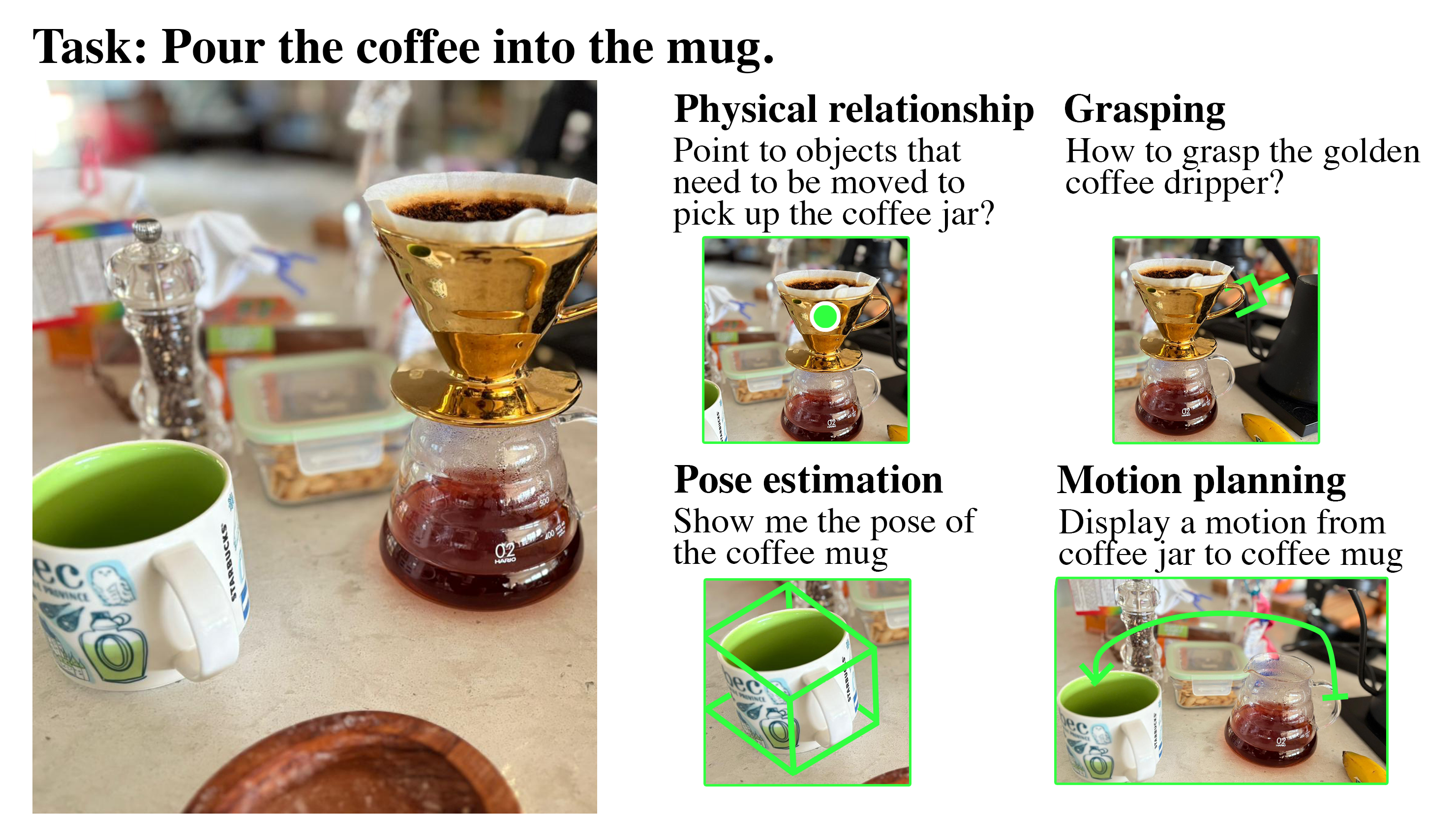}
    \caption{
    The \datasetname{} dataset facilitates 
    object-interaction reasoning
    for robot manipulation. 
    This illustration demonstrates how a model trained on \datasetname{} enables 
    human and robot-aligned spatial understanding for different actions, supporting physical relationship, 
    locating where to grasp objects, 
    precise pose estimation, and 
    motion planning between objects.
    }
    \label{fig:teaser}
\end{figure}

Vision-language models (VLMs) have demonstrated impressive performance across robotics, augmented reality, and egocentric vision tasks, 
supporting capabilities such as scene description~\cite{fangandliu2024moka}, 
or code generation for robot control~\cite{singh2023progprompt, codeaspolicies2022}. 
Despite this progress, further improvements are needed to enable embodied applications of VLMs. For example, they may correctly identify the coffee jar in a cluttered scene (see Figure~\ref{fig:teaser}), but struggle to determine exactly where to grasp it, 
how to navigate around  objects, 
or which items must be moved first to access it, capabilities essential for real-world embodiment. 
We define this gap as \textit{object-interaction reasoning}: 
the ability to understand and predict fine-grained physical relationships between objects, 
including grasp affordances, collision-aware motion paths, and manipulation sequencing in cluttered environments.

\begin{table*}[t]
    \centering
    \small
    \renewcommand\arraystretch{0.9} 
    \captionsetup{width=\textwidth} 
    \caption{Comparison with spatial reasoning datasets including reference frames and whether they provide motions, poses, and grasping. Only \datasetname{} includes all three.}
    \resizebox{\textwidth}{!}{%
    \begin{tabular}{c|ccccccc} 
        \toprule
        \textbf{Dataset} & \textbf{Domain} & \textbf{Ref. Frames} & \textbf{\# Images} & \textbf{\# Spatial Q\&As} & \textbf{Motions} & \textbf{Poses} & \textbf{Grasping} \\
        \midrule 
        EmbSpatial-Bench~\cite{du-etal-2024-embspatial}   & Indoor            & \xmark & \,\,2k   & \,\,4k   & \xmark & \xmark & \xmark \\
        Visual Spatial~\cite{liu-etal-2023-visualspatial} & MSCOCO            & \cmark & 10k      & 10k      & \xmark & \xmark & \xmark \\
        SpatialRGPT-Bench~\cite{cheng2024spatialrgpt}     & Indoor, AV        & \xmark & 1.4k     & 1.4k     & \xmark & \xmark & \xmark \\
        BLINK-Spatial~\cite{fu2024blink}                  & Generic           & \cmark & 286      & 286      & \xmark & \xmark & \xmark \\
        What's up~\cite{kamath2023whatsup}                & Generic           & \xmark & \,\,5k   & 10k      & \xmark & \xmark & \xmark \\    
        Spatial-MM~\cite{shiri-etal-2024-spatialmm}       & Generic           & \cmark & 2.3k     & 2.3k     & \xmark & \xmark & \xmark \\
        RoboSpatial~\cite{song2025robospatial}            & Indoor, tabletop  & \cmark & \,\,1M   & \,\,3M   & \xmark & \xmark & \xmark \\
        RoboBrain~\cite{ji2025robobrain}                  & Tabletop          & \xmark & \,\,667k   & \,\,1M   & \cmark & \xmark & \xmark \\
        RoboRefer~\cite{zhou2025roboreferspatialreferringreasoning} 
                                                          & Mixed  & \xmark & 550k & 20M & \xmark & \xmark & \xmark
        \\
        \midrule
        \datasetname                                      & Tabletop          & \cmark & 150k      & 33.8M      & \cmark & \cmark & \cmark \\
        \bottomrule 
    \end{tabular}%
    }
    
    \label{tab:dataset_comparison}
\end{table*}

Recent efforts such as SpatialVLM~\cite{chen2024spatialvlm} and RoboPoint~\cite{yuan2024robopoint} have taken steps toward training VLMs with explicit spatial objectives. 
While these approaches improve performance on synthetic or internet-scale benchmarks, 
they often under-perform with object-to-object relationships, manipulation or long-term reasoning questions.
More recently, RoboSpatial~\cite{song2025robospatial} introduced a spatial reasoning benchmark, 
establishing an important first step toward bridging perception, spatial reasoning and spatial relationships. 
In parallel, other approaches such as MolmoAct~\cite{lee2025molmoact} and RoboBrain~\cite{ji2025robobrain} train models directly on robot datasets to predict manipulation trajectories and motions. 
While these directions are promising, they remain limited in scope, especially in object-interaction reasoning: they do not capture the breadth of fine-grained physical interactions needed for general reasoning, 
such as grasp feasibility, pose estimation, physical relationships, 
object-based motion planning.

To bridge this gap, we introduce \datasetname, a new dataset that extends BOP (Benchmark for Object Pose estimation)~\cite{nguyen2025bop} into the domain of object-interaction reasoning. 
While BOP traditionally provides high-quality 3D annotated images with accompanying 3D models 
for object detection and pose recovery, it does not directly address reasoning about interactions between objects. 
In contrast to prior spatial reasoning datasets that rely on approximate methods for annotation~\cite{chen2024spatialvlm}---such as monocular depth estimation---\datasetname{} inherits precise 3D ground-truth poses from BOP, enabling reasoning at a level of accuracy unseen in the spatial reasoning datasets. 
\datasetname{} introduces a diverse set of question-answer pairs targeting 
object-interaction reasoning, such as motion planning, physical relationships, pose estimation, grasping, \textit{etc.} 
The stated goals for our dataset are: 
1) Precision, in contrast to existing datasets that use approximate annotations that cannot support the millimeter-level accuracy required for grasping and motion planning~\cite{chen2024spatialvlm,fu2024blink};
2) Interaction completeness to capture the full chain of reasoning from perception to executable manipulation without the need of multiple choice questions, overcoming the limitations of current benchmarks~\cite{song2025robospatial,du-etal-2024-embspatial,tong2024cambrian1} that
only evaluate spatial relationships, and, 
3) Scale and diversity: spatial reasoning datasets remain small, while large-scale datasets lack tasks related to object interaction~\cite{yuan2024robopoint}.
Finally \datasetname{} uses pixel-level question-answers where models are asked to output pixel locations with great precision, this is a departure from previous spatial reasoning datasets that focused on 
multiple-choice or yes/no questions~\cite{geminiER15,fu2024blink,chen2024spatialvlm,song2025robospatial}.

\datasetname{} is designed to serve as both a large-scale training corpus and a benchmark for evaluating object-interaction reasoning. 
The training dataset consists of approximately 150k high-quality images drawn from the BOP benchmark, from which we  generate over 33M question–answer pairs spanning object relationships, manipulation affordances, motion feasibility, and scene-level reasoning.
This scale already surpasses prior spatial reasoning datasets, while maintaining accurate 6D pose annotations as geometric grounding (see Table~\ref{tab:dataset_comparison}). 

Accompanying the training data, we also introduce two new evaluation benchmarks: \datasetname-core is formed from hand-selected held-out BOP scenes, and \datasetname-lab is hand-constructed from images taken in different labs,
ensuring that it is fully independent of BOP, \textit{e.g.}, camera, object, point of view.  
Evaluations of both open- and closed-source vision–language models on our hand-curated benchmarks (core and lab) show that existing systems struggle with object-interaction reasoning, underscoring the need for datasets that more directly probe geometric grounding. 
These results are also striking when we compare them to the results obtained from human evaluation of the same tasks in \datasetname-core as these models.  
We also observe clear gains in real-world robot performance when using the model fine-tuned on \datasetname.
We highlight our contributions as follows: 
\begin{itemize}[noitemsep, topsep=0pt, parsep=0pt]
    \item We introduce a novel object-interaction reasoning dataset built on BOP, 
    augmenting perception data with queries about object interactions, 
    manipulation affordances, 
    motion feasibility, and scene-level reasoning. 
    \item We show that including \datasetname{} into the training recipe improves VLM performance not only on our test sets but also on out-of-domain spatial reasoning benchmarks such as 
    RoboSpatial-Home~\cite{song2025robospatial}, 
    SpatialBench~\cite{du-etal-2024-embspatial}, 
    and CV-Bench~\cite{tong2024cambrian1}.
    \item We propose \datasetname-core and \datasetname-lab as benchmarks for embodied spatial and object-interaction reasoning, bridging the gap between pixel-level perception and high-level reasoning. 
\end{itemize}

\section{Related Work}

\noindent\textbf{Spatial reasoning.} 
Spatial reasoning has long been an implicit and explicit component of numerous vision and question answering tasks~\cite{fu2024blink, scanqa, jia2024sceneversescaling3dvisionlanguage, suhr2019nlvr2visualbiasanalysis, salewski2022clevrx, krishna2017visualgenome, johnson2017clevr, hudson2019gqa,zhao2025manipbench}. 
While many benchmarks and methods have been proposed, they show limitations: some are restricted to simulations~\cite{szymanska2024space3dbench} or generic image datasets~\cite{liu-etal-2023-visualspatial, rajabi2024groundedvisualspatialreasoning, cheng2024spatialrgpt, chen2024spatialvlm, fu2024blink, kamath2023whatsup, shiri-etal-2024-spatialmm, ranasinghe_spatial}, others are challenging to evaluate due to reliance on free-form text outputs~\cite{szymanska2024space3dbench, du-etal-2024-embspatial, linghu2024multi-modalsituated}, 
some require complete 3D scans~\cite{zhang2024spartun3dsituatedspatialunderstanding, man2024situation3d, ma2022sqa3d, linghu2024multi-modalsituated}, 
and many fail to incorporate 
object-interaction reasoning or robotics~\cite{zhang2024spartun3dsituatedspatialunderstanding, man2024situation3d, ma2022sqa3d, linghu2024multi-modalsituated, chen2024spatialvlm, cheng2024spatialrgpt, fu2024blink, ranasinghe_spatial}. 
Moreover, several works overlook actionable, robotics-relevant spatial relationships such as spatial compatibility, grasping, motion, and contextual reasoning~\cite{du-etal-2024-embspatial, wang2023embodiedscan, shiri-etal-2024-spatialmm, kamath2023whatsup, linghu2024multi-modalsituated, ranasinghe_spatial,ray2025satdynamicspatialaptitude}.

\noindent\textbf{VLMs for robotics.} 
VLMs have become pivotal in robotics, enabling systems to interpret and act upon complex visual and textual inputs~\cite{singh2023progprompt,chenrobo2vlm}. 
By integrating visual perception with natural language understanding, VLMs enable more intuitive human–robot interaction and strengthen autonomous decision-making. 
Recent progress has demonstrated their potential across diverse robotic applications. 
For example, vision-language-action models (VLAs)~\cite{kim24openvla, rt22023arxiv, octo_2023} allow robots to parse complex instructions and output executable actions. 
VLMs like GPT-4v~\cite{openai2022chatgpt} have been leveraged for high-level task planning~\cite{wake2023gpt}, generating action sequences directly from natural language prompts. 
Other applications include keypoint/mask prediction~\cite{huang2024rekep, wi2023calamari, google2024pivot,zhou2025roboreferspatialreferringreasoning}, error analysis~\cite{duan2024ahavisionlanguagemodeldetectingreasoning, Song_2022_CVPR}, and grasp pose estimation~\cite{huang2024copageneralroboticmanipulation}. 
Recent efforts have explored grasping and motion planning: MolmoAct~\cite{lee2025molmoact} learns from robot trajectories to generate executable actions, while SpatialPin~\cite{ma2024spatialpin} uses a planner to generate motions, though the model itself cannot output motion sequences directly. 

%
More closely related, prior studies on spatial reasoning~\cite{liu-etal-2023-visualspatial, kamath2023whatsup} and RoboSpatial~\cite{song2025robospatial} introduced datasets for reasoning about object reference frames, showing strong potential for robotic applications and spatial reasoning research.
Inspired by Robospatial, we further extend the set of tasks that model can now tackle such as grasping, motion planning, object rearrangement, and pose estimation.
We note the concurrent work of TIGeR~\cite{han2025tiger}, which similarly proposes a dataset for object-interaction reasoning, encompassing tasks like localization, distance estimation, and pose computation. As their dataset and models were not publicly available at the time of writing, a direct comparison on shared tasks was not feasible.





\section{Methodology}

In this section, we introduce \textit{Object-Interaction Reasoning}—a set of perception and reasoning skills essential for deploying VLMs as embodied agents in real-world robotic environments. We first outline the six core skills that constitute this reasoning framework and discuss their relevance to physical manipulation. We then describe our automated data generation pipeline for constructing \datasetname{}, a large-scale, geometrically grounded dataset for training and evaluating these capabilities.

\subsection{Object-Interaction Reasoning}

We identify six fundamental skills that a VLM needs to serve as a robust perceptual interface for embodied reasoning and manipulation.  \textit{(1) Object pose estimation} involves predicting the 3D bounding box of a referred object, enabling precise visual grounding in a scene. This extends beyond conventional VLMs, which typically mark only points or 2D rectangles.  \textit{(2) Object grasp estimation} trains the model to infer stable 3D grasp poses for target objects, providing actionable inputs for downstream robotic execution and manipulation tasks.  \textit{(3) Inter-object motion prediction} requires generating collision-free waypoints to move a source object toward a target object. This skill integrates spatial awareness and motion planning, allowing the model to reason about dynamic object-pair interactions in cluttered environments.  \textit{(4) Object rearrangement} assesses the ability to understand severely cluttered workspaces and identify which obstructing objects must be moved before grasping a target.  This skill allows the model to declutter the workspace before attempting to grasp a target object. Finally, following prior works, we include two binary reasoning tasks: \textit{(5) spatial reasoning} and \textit{(6) relative depth perception}. The former focuses on identifying spatial relationships such as ``left of,'' ``right of,'' ``above,'' and ``below,'' while the latter determines relative distance attributes (``closer'' and ``farther'') between object pairs.  Together, these six skills provide a unified framework for object-centric perception and reasoning. A VLM finetuned under this formulation gains general-purpose scene understanding and manipulation abilities, with its intrinsic pre-training enabling improved zero-shot transfer to unseen objects and environments.

\subsection{Dataset Generation}

\begin{figure}[ht]
    \centering
    \includegraphics[width=1.0\linewidth]{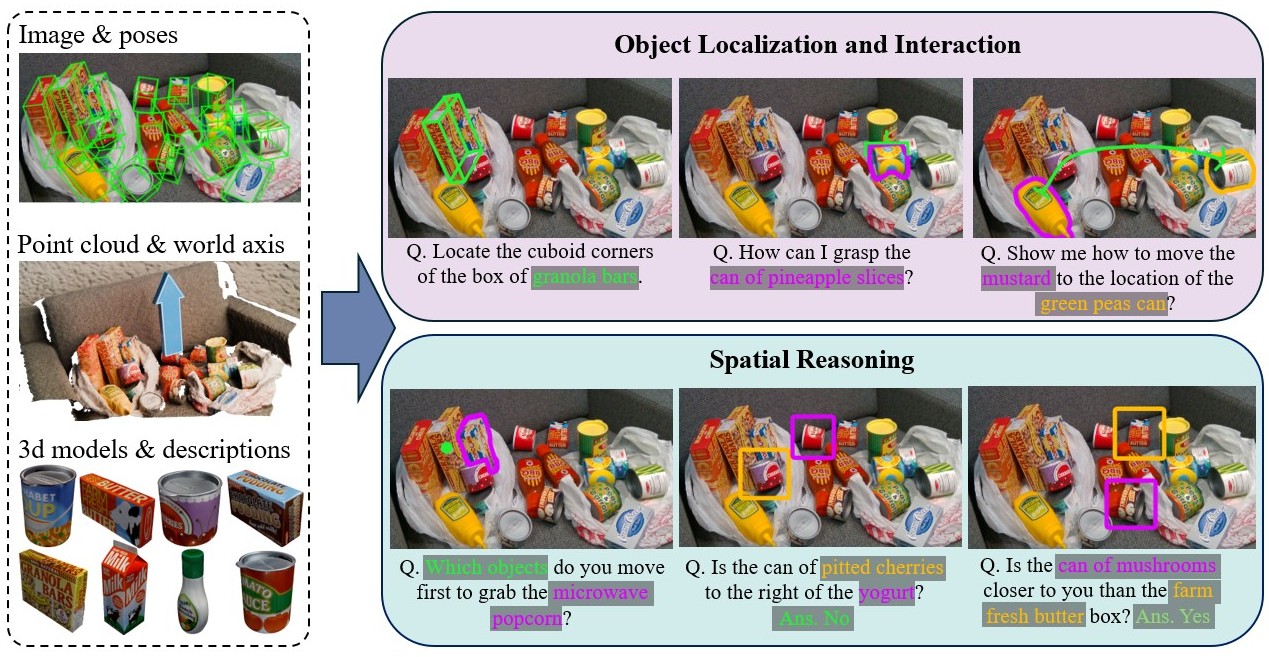}
    \vspace{-6mm}
    \caption{
    Overview of the \datasetname{} dataset. 
    We automatically generate object-interaction and spatial reasoning annotations 
    from 3D point clouds, images, object poses and 3D models with description. 
    We create question/answer pairs covering 6 types of questions (from left to right, top to bottom),
    object pose estimation, grasp affordance, motion planning, physical interaction, object relationship, and 
    depth relationship. 
    }
    \label{fig:data-gen}
\end{figure}

Our data generation framework is designed to produce rich, geometrically consistent interaction data with minimal manual supervision. It grounds the dataset in accurate 3D object poses while automatically generating diverse, object-centric metadata describing interactions and spatial relationships.  
Formally, the framework takes as input a pose dataset $D_p$ containing RGB-D images, 
3D object poses, camera intrinsics, 
and outputs an object-interaction reasoning dataset where each sample $S = \langle I_k, Q_k, A_k, T_k \rangle$ comprises the input image $I_k$, a question $Q_k$, its corresponding answer $A_k$, and a task label $T_k$.  Figure~\ref{fig:data-gen} describes our dataset pipeline. 

\subsubsection{World frame construction}  
A typical scene in a pose estimation dataset consists of RGB-D images, camera intrinsics, and ground-truth object poses. Estimating the world-frame $Z$ direction from object poses alone is unreliable, as the orientation of cluttered objects offers a poor proxy for gravity. To reconstruct a consistent world coordinate system, we estimate the camera-to-world transformation $^{cam}T_{world} \in SE(3)$.  
%
%
We first localize the planar support surface (\textit{e.g.}, a table) on which objects rest using a pointing focused VLM~\cite{deitke2024molmopixmoopenweights} 
and fit a plane to the corresponding 3D points via RANSAC. 
The plane normal $\mathbf{n}_{p}$ defines the world up direction. Let $\mathbf{v}_{z} = [0,0,1]^{\top}$ denote the canonical world up-axis. 
The rotation aligning $\mathbf{v}_{z}$ to $\mathbf{n}_{p}$ is computed using the Rodrigues rotation formula~\cite{rodrigues1840lois}.
Finally, the translation vector $\mathbf{t}$ is determined such that the fitted plane aligns with the world origin, ensuring all scene coordinates are expressed in the same frame.

\subsubsection{Geometrical priors}  
Our data generation algorithm produces a physically grounded, geometry-aware dataset from a rich source of information: object 3D poses. 
First we compute the 3D cuboid bounding boxes from each scene using the object pose and model sizes. Motion trajectories are synthesized using a Rapidly-exploring Random Tree (RRT) planner~\cite{lavalle1999rrt} operating in 3D Cartesian space to compute collision-free pick-and-place paths between object locations. 
For $n$ objects in a scene, we generate $\tbinom{n}{2}$ inter-object trajectories by initializing the planner at one object’s centroid and applying a 10\% goal-biased random sampling toward the second object. Paths that intersect with neighboring object meshes—derived from their 3D poses—are filtered out.  The raw RRT paths are then refined using the Ramer--Douglas--Peucker algorithm~\cite{douglas1973algorithms} for 3D path simplification, reducing redundant waypoints and producing smooth, physically plausible trajectories. 

Object grasps are computed in the reconstructed world frame using the transformer-based parallel gripper model M2T2~\cite{yuan2023m2t2}, which employs a dual-sampling strategy combining global scene points for contextual reasoning and object-centric points for accurate geometric representation. We retain the top-5 grasps per object to ensure diversity in contact configurations.  If all predicted grasps for an object result in collisions with surrounding objects, these are discarded and the object is labeled as \textit{fully cluttered}. This labeling forms the foundation of the \textit{object rearrangement} skill, where the VLM must identify which objects should be moved first to enable a successful grasp. 
Since the data source we are injecting includes multiple camera viewpoints, the resulting combination of poses, grasps, and motion trajectories yields a densely annotated, geometrically consistent dataset. 

\subsubsection{Generating Question-Answer pairs}  

Once the geometrical priors are generated, we form question and answer pairs using object description and templates.
First, we render the 3D models used in the scene, these renders are sent to a VLM to acquire object descriptions including: shape, color, size, and general utility. 
Each description is then manually verified and refined when necessary to ensure accuracy and linguistic diversity. 
The metadata generated in Section~3.2.2 serves as input for constructing VQA templates. We design a distinct template for each reasoning task, following the structure \texttt{\{TASK\_TYPE\} \{OBJECT~A\} \{OBJECT~B\}}. These templates are then supplied to the LLM, together with carefully curated in-context examples, to generate linguistically diverse and human-like questions.
%
%
%
Robotic scenes often contain multiple instances of the same object category, where visual cues such as color, shape, or texture alone cannot uniquely disambiguate the target.
To distinguish multiple instances of the same object category within a scene, 
we compute the centers of their 3D bounding boxes and assign relative positional attributes \texttt{\{POS.\ ATTRIBUTE\}} such as ``leftmost'', ``rightmost'', ``topmost'', and ``bottommost''. These attributes are then concatenated with the above question generation template for objects that have multiple instances in a particular scene.
%
%
%

Object poses, grasps, trajectories, and rearrangements are represented as ordered 2D lists of keypoints. 
Spatial reasoning and relative depth perception tasks are annotated as binary (``yes''/``no'') responses.  
The final dataset comprises approximately 33M high-quality VQA pairs spanning the six object-interaction reasoning skills.

\begin{table}[ht]
\centering
\small
\renewcommand{\arraystretch}{0.85}  
\caption{Distribution of questions and frequency in \datasetname}
\label{tab:qtype-distribution}
\begin{tabular}{lrr}
\toprule
\textbf{Question Type} & \textbf{\# Q\&As} & \textbf{\% of Total} \\  
\midrule
Object Poses                  & 4.2M  & 12.4\% \\
Grasps                        & 5.4M  & 16.0\% \\
Trajectories                  & 5.4M  & 16.0\% \\
Object rearrangement          & 2.6M  & 7.6\%  \\
Spatial Reasoning             & 5.4M  & 16.0\% \\
Relative Depth Perception     & 10.8M & 32.0\% \\
\midrule
Total                         & 33.8M & 100\%  \\
\bottomrule
\end{tabular}
\end{table}



\section{\datasetname: A Spatial Reasoning Benchmark for Object-Interaction Reasoning}
%
We choose the BOP~\cite{nguyen2025bop} family of datasets which provide both real and simulated images comprising diverse table-top scenes in indoor and outdoor environments. 
Our data generation pipeline results in three main datasets: a training dataset (\datasetname) and two test datasets: (\datasetname-core and \datasetname-lab).
%
%
In this section, we present the three datasets with their accompanying task metrics.  

\paragraph{\datasetname}
We utilize images and pose annotations from HOPE~\cite{tyree2022hope}, HANDAL~\cite{handal}, YCB-V~\cite{posecnn}, 
and LineMOD~\cite{Hinterstoier2012ModelBT} as our base source for generating rich spatio-geometric visual question-answers (VQAs). 
These datasets were chosen to capture sufficient scene and object diversity, variations in camera poses and clutter as observed in real-world robot environments. 
Our datasets span high-clutter and geometric variability (HOPE), 
increased object diversity (HANDAL), 
and lower-resolution but varied scenes (YCB-V and LineMOD), 
providing a broad range of visual conditions.


    
    



Our final dataset includes 104 unique household and industrial objects arranged in high degrees of clutter and dense environments. 
Many objects contain textual labels such as ``Milk" or ``BBQ Sauce" which make them easier to identify when viewed from the front. However, occlusions and randomized arrangements demand global scene understanding and referring attributes such as color and position to uniquely identify the objects. 
Some scenes also contain distractors (multiple instances of the same category) making visual grounding challenging which affects subsequent grasping and trajectory predictions. 
After rigorous manual (established from different dataset metrics) and occlusion filtering for grasps and trajectories, 
our dataset contains 150K unique images and 33M QA pairs. 
Table~\ref{tab:qtype-distribution} summarizes the distribution of question types across the dataset. 
While the data pipeline prioritizes extracting maximum number of high-quality spatio-geometric annotations for adapting VLMs for spatial indoor tasks, 
our algorithm can be easily extended to novel scenes with object poses, which is trivial to obtain using simulated environments. 


\paragraph{\datasetname{} Test sets. }
We release \datasetname-core and \datasetname-lab, two test corpora comprising high-quality VQA pairs. 
Specifically for \datasetname-core we use held-out scenes from BOP and each VQA pair is manually verified using a semi-automated pipeline where the annotations created by the framework are manually updated if necessary. In total,  we release 688 VQA pairs as a part of our testing benchmark, with the following distribution: object poses (17.4$\%$), object grasps (17.4$\%$), trajectories (17.4$\%$), spatial reasoning (17.4$\%$), relative depth perception (23.3$\%$) and object rearrangement (7.1$\%$). 
We also release \datasetname{}-lab with images not present in BOP datasets to measure the performance on out-of-domain objects. This corpus is manually annotated and comprises 240 VQA pairs from 15 images. 

\begin{table*}[!t]
\centering
\setlength{\tabcolsep}{5pt}
\small
\renewcommand{\arraystretch}{0.95}
\caption{Performance of popular VLMs on \datasetname-core, 
which features cluttered object configurations requiring fine-grained spatio-geometric reasoning for robotics. Note that `inf' indicates that the model did not produce a valid output. Bold and underlined scores indicate the best and second best performing methods for a particular task.
}
\label{tab:mainresults}
\begin{tabular}{l
                c
                cc
                c
                c
                c
                c}
\toprule
\multirow{2}{*}{\textbf{Method}} &
\multicolumn{1}{c}{\textbf{Pose}} &
\multicolumn{2}{c}{\textbf{Trajectory}} &
\multicolumn{1}{c}{\textbf{Grasps}} &
\multicolumn{1}{c}{\textbf{Spatial}} &
\multicolumn{1}{c}{\textbf{Rel. Depth}} &
\multicolumn{1}{c}{\textbf{Obj. Rearr.}} \\
\cmidrule(lr){2-2}\cmidrule(lr){3-4}\cmidrule(lr){5-5}\cmidrule(lr){6-6}\cmidrule(lr){7-7}\cmidrule(lr){8-8}
 & \textbf{3D IOU $\uparrow$} & \textbf{SR $\uparrow$} & \textbf{Dist. Err. $\downarrow$} & \textbf{NCE $\downarrow$} & \textbf{SR $\uparrow$} & \textbf{SR $\uparrow$} & \textbf{Recall (\%) $\uparrow$} \\
\midrule
Human & 54.2 & \textbf{67.3} & 112.4 & \textbf{1.1} & 84.9 & 87.3 & 44.1 \\
\midrule
\multicolumn{8}{c}{\textbf{Proprietary Models}} \\
\midrule
GPT-5~\cite{GPT5} & 9.0 & 0 & inf & inf & 68.3 & 74.6 & 14.8 \\
Gemini Robotics-ER 1.5~\cite{geminiER15} & 24.4 & 43.0 & 138.4 & 4.2 & 84.2 & 88.0 & 48.9 \\
\midrule
\multicolumn{8}{c}{\textbf{Open-Source Models}} \\
\midrule
Molmo~\cite{deitke2024molmopixmoopenweights} (72B) & 22.6 & 13.2 & 224.8 & \underline{1.4} & 68.3 & 65.6 & 28.0 \\
Qwen-VL 2.5~\cite{Qwen2.5-VL} (3B) & 26.5 & 0 & 240.6 & 3.1 & 50.8 & 49.4 & 15.2 \\
NVILA~\cite{Liu_2025_CVPR} (2B) & 6.5 & 0 & inf & 8.2 & 65.0 & 52.5 & 8.3 \\
NVILA~\cite{Liu_2025_CVPR} (15B) & 27.2 & 6.2 & 206.3 & 5.3 & 75.0 & 65.0 & 25.0 \\
RoboRefer~\cite{zhou2025roboreferspatialreferringreasoning} & 34.3 & 0 & inf & inf & 81.7 & 84.0 & 16.6 \\
Intern-VL~\cite{chen2024internvl} & 18.3 & 0 & inf & inf & 72.5 & 76.0 & 8.6 \\
\midrule
\multicolumn{8}{c}{\textbf{Models Trained on \datasetname}} \\
\midrule
Qwen-VL 2.5 (3B) - SFT & 48.2 & 22.5 & 116.3 & 1.5 & 92.6 & \underline{94.1} & 43.4 \\
NVILA (2B) - SFT & \textbf{77.4} & 50.8 & \underline{78.5} & 1.69 & \underline{94.2} & \textbf{94.6} & \underline{56.4} \\
NVILA (15B) - SFT &\underline{73.5} & \underline{64.2} & \textbf{77.4} & 1.40 & \textbf{95.8} & \textbf{94.6} & \textbf{57.7} \\
\bottomrule
\end{tabular}
\end{table*}



\paragraph{Evaluation Metrics.}  
We adopt standard 3D metrics and success rates to evaluate model performance across all spatial reasoning tasks in \datasetname{}.  
For \textit{object pose estimation}, we report the Intersection-over-Union (IoU) metric, which measures the volumetric overlap between the 3D projected cuboids of the predicted and ground-truth poses.  
For \textit{trajectory prediction}, we employ two complementary metrics: (i) \textit{success rate}, defined as $1$ if the initial and final points of the predicted trajectory lie on the corresponding source and target objects, and (ii) \textit{distance error}, computed as the mean per-pixel distance between predicted and ground-truth waypoints. The success rate evaluates whether the model correctly identifies the relevant object pair for manipulation, while the distance error quantifies the geometric accuracy of the predicted trajectory. While multiple valid motion paths may exist between two objects, distance error provides a continuous measure of how closely the predicted trajectory aligns with the ground-truth path.  
For \textit{object rearrangement}, we report Recall (\%)—the proportion of correctly identified obstructing objects that must be moved relative to the total number of required moves. Finally, for \textit{spatial reasoning} and \textit{relative depth perception}, we compute the average success rate over all VQA pairs.


Samples in \datasetname{} adopt a grasp representation distinct from conventional 2D grasp datasets. We employ a five-point formulation that specifies the grasp center, left finger base, right finger base, left finger tip, and right finger tip. This representation captures both position and orientation more comprehensively than rectangle-based~\cite{11128679,bhat2025maplegrasp} or point-based affordance formulations~\cite{deshpande2025graspmolmo}, offering a more flexible transition to 3D by allowing greater freedom in selecting the grasp plane.  
To evaluate grasp prediction performance, we use the \textit{normalized coordinate error} (NCE) metric, which measures the mean normalized Euclidean distance between corresponding predicted and ground-truth grasp points:
\[
\text{NCE} = \frac{1}{N} \sum_{i=1}^{N} \frac{\| p_i - \hat{p}_i \|_2}{d},
\]
where $N=5$, $p_i$ and $\hat{p}_i$ denote the predicted and ground-truth grasp points, respectively, and $d$ is the gripper width. Each point is normalized by the image width and height to ensure scale invariance.

\begin{table*}[ht]
    \centering

    \begin{tabular}{m{0.045\textwidth} p{0.93\textwidth}}
        \rotatebox{90}{\textbf{Pose}} &
        \begin{minipage}{\linewidth}\centering
            \begin{minipage}{0.3\linewidth}\centering
                \qimg{Draw the cuboid for the orange spatula?}{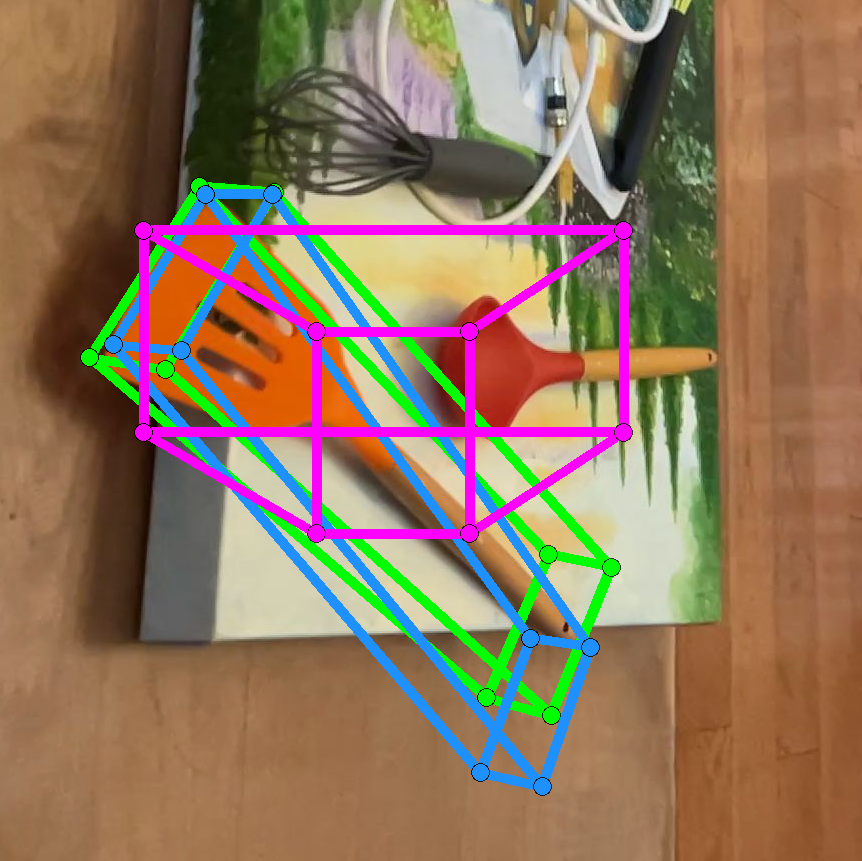}
            \end{minipage}\hspace{0.5em}
            \begin{minipage}{0.3\linewidth}\centering
                \qimg{Draw the cuboid of the rightmost butter box?}{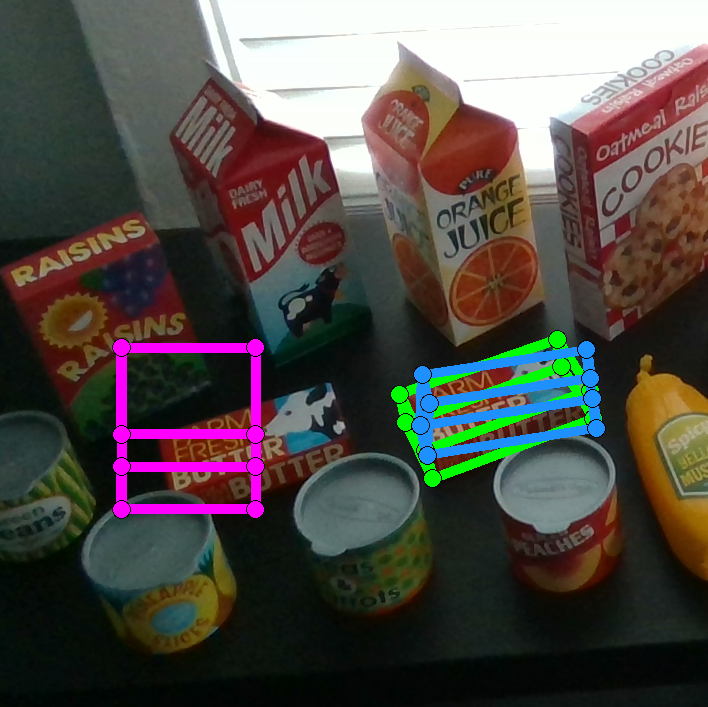}
            \end{minipage}\hspace{0.5em}
            \begin{minipage}{0.3\linewidth}\centering
                \qimg{Draw the cuboid of the Cheez-it box?}{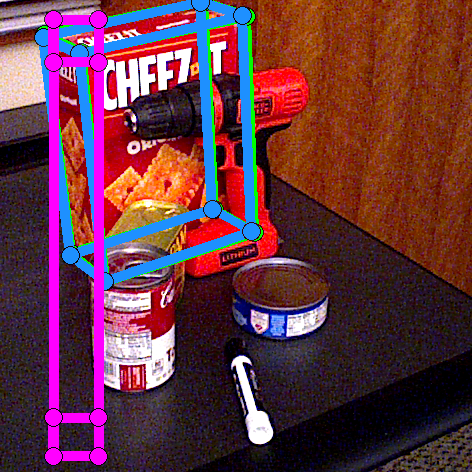}
            \end{minipage}
        \end{minipage} \\\hline\noalign{\vskip 4pt}
        \rotatebox{90}{\textbf{Grasp}} &
        \begin{minipage}{\linewidth}\centering
            \begin{minipage}{0.3\linewidth}\centering
                \qimg{Draw the grasp for the pineapple can?}{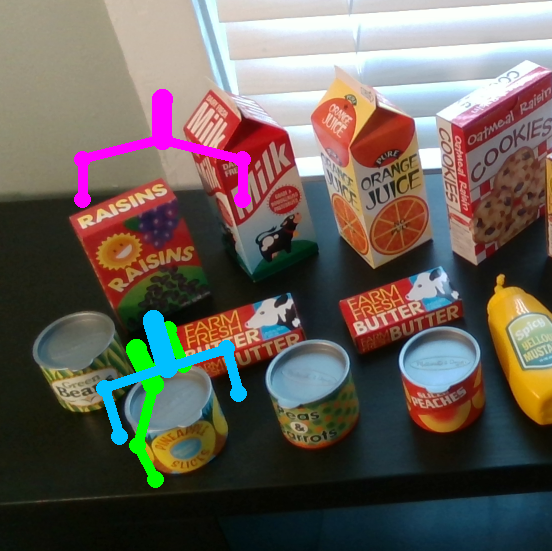}
            \end{minipage}\hspace{0.5em}
            \begin{minipage}{0.3\linewidth}\centering
                \qimg{Draw the grasp for the heaphones? (lab)}{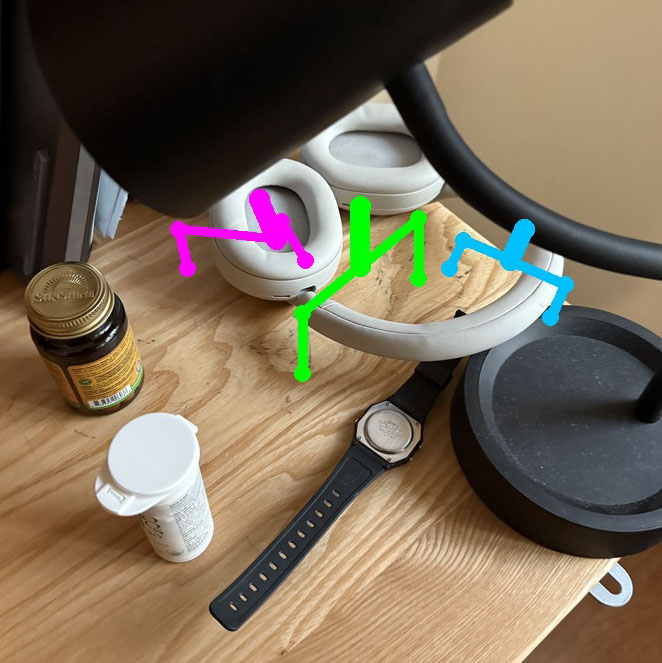}
            \end{minipage}\hspace{0.5em}
            \begin{minipage}{0.3\linewidth}\centering
                \qimg{Draw the grasp of shiny bowl?}{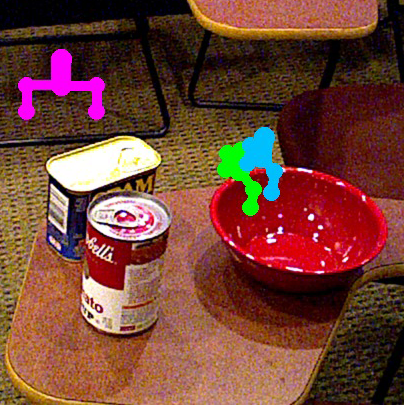}
            \end{minipage}
        \end{minipage} \\\hline\noalign{\vskip 4pt}
        \rotatebox{90}{\textbf{Trajectory}} &
        \begin{minipage}{\linewidth}\centering
            \begin{minipage}{0.3\linewidth}\centering
                \qimg{What is the path from water bottle to the tennis ball? (lab)}{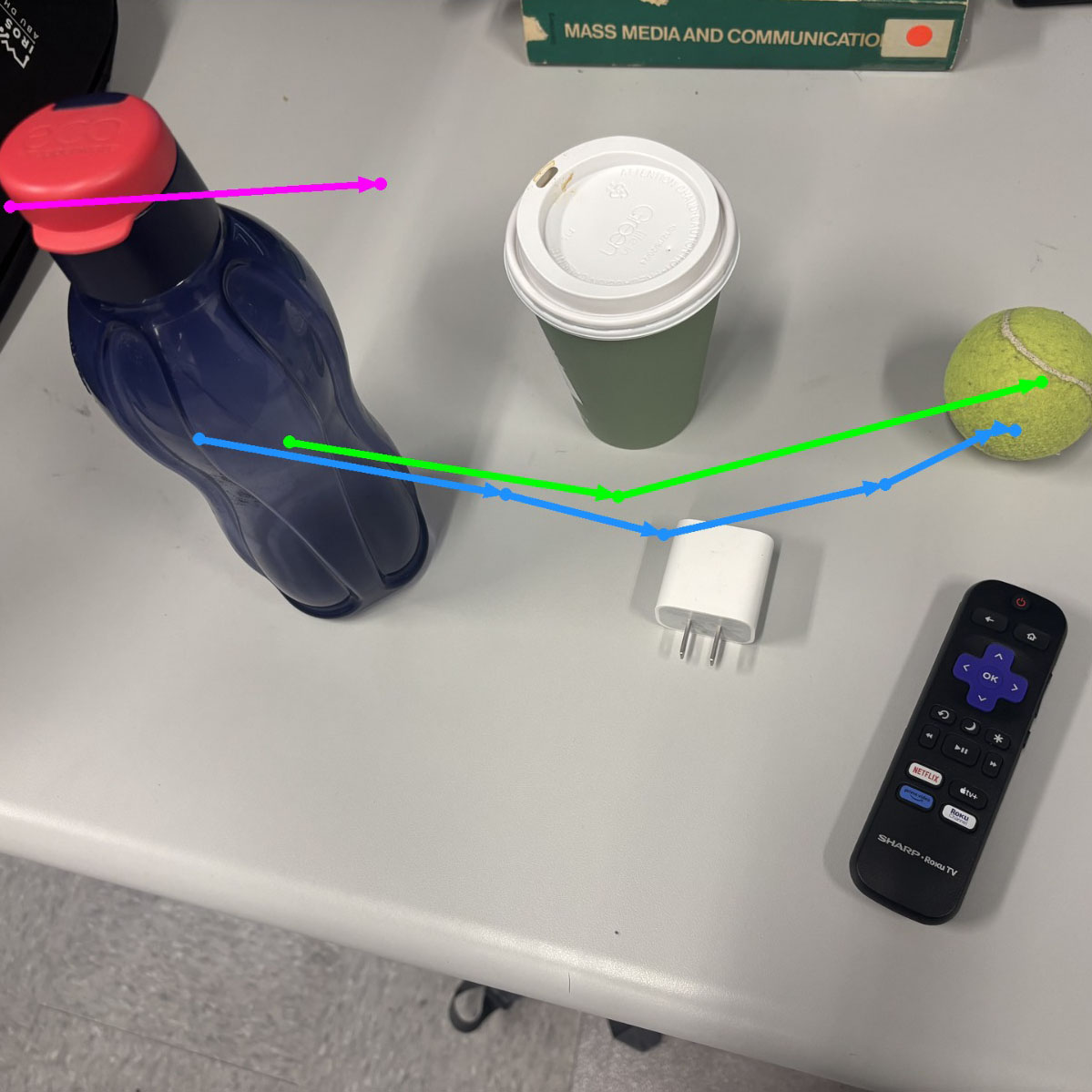}
            \end{minipage}\hspace{0.5em}
            \begin{minipage}{0.3\linewidth}\centering
                \qimg{What is the path from the whisk to rounded colander?}{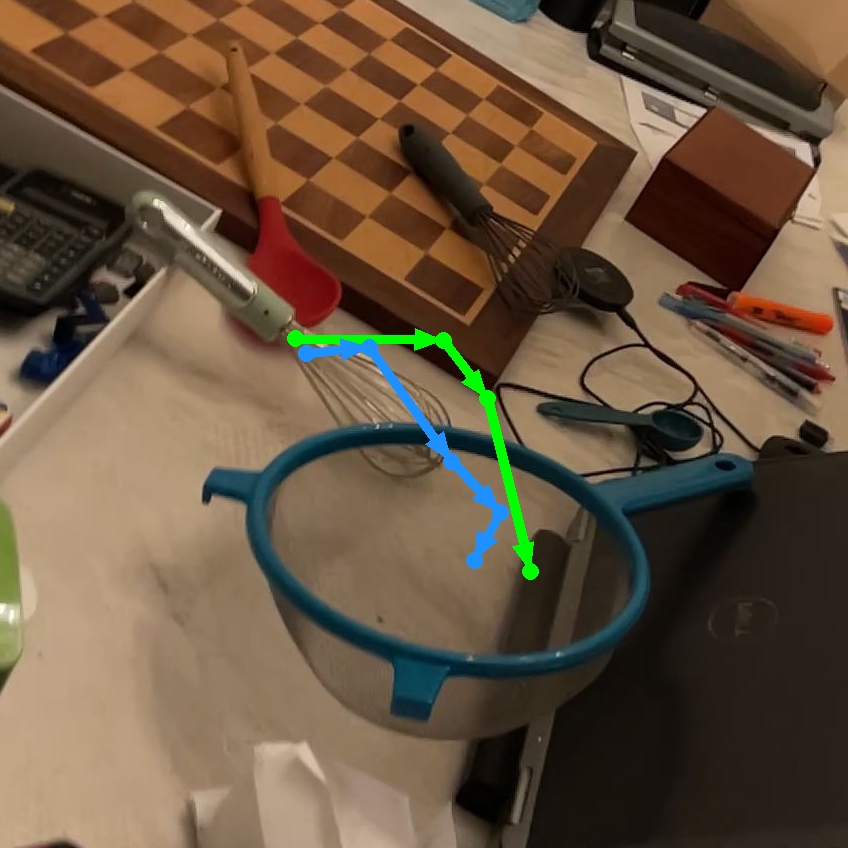}
            \end{minipage}\hspace{0.5em}
            \begin{minipage}{0.3\linewidth}\centering
                \qimg{What is the path from the granola bar to the cherry can?}{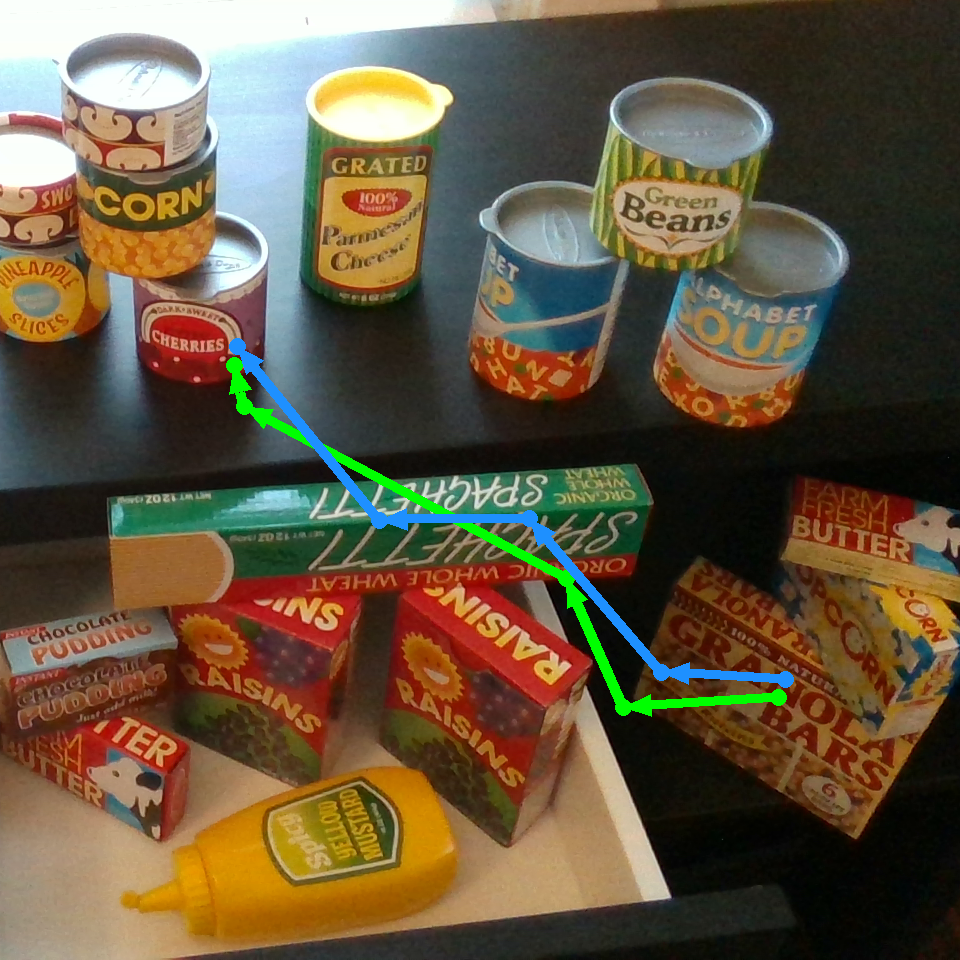}
            \end{minipage}
        \end{minipage} \\\hline\noalign{\vskip 4pt}
        \rotatebox{90}{\textbf{Rearrangement}} &
        \begin{minipage}{\linewidth}\centering
            \begin{minipage}{0.3\linewidth}\centering
                \qimg{Which object do you need to remove to grab the laptop? (lab)}{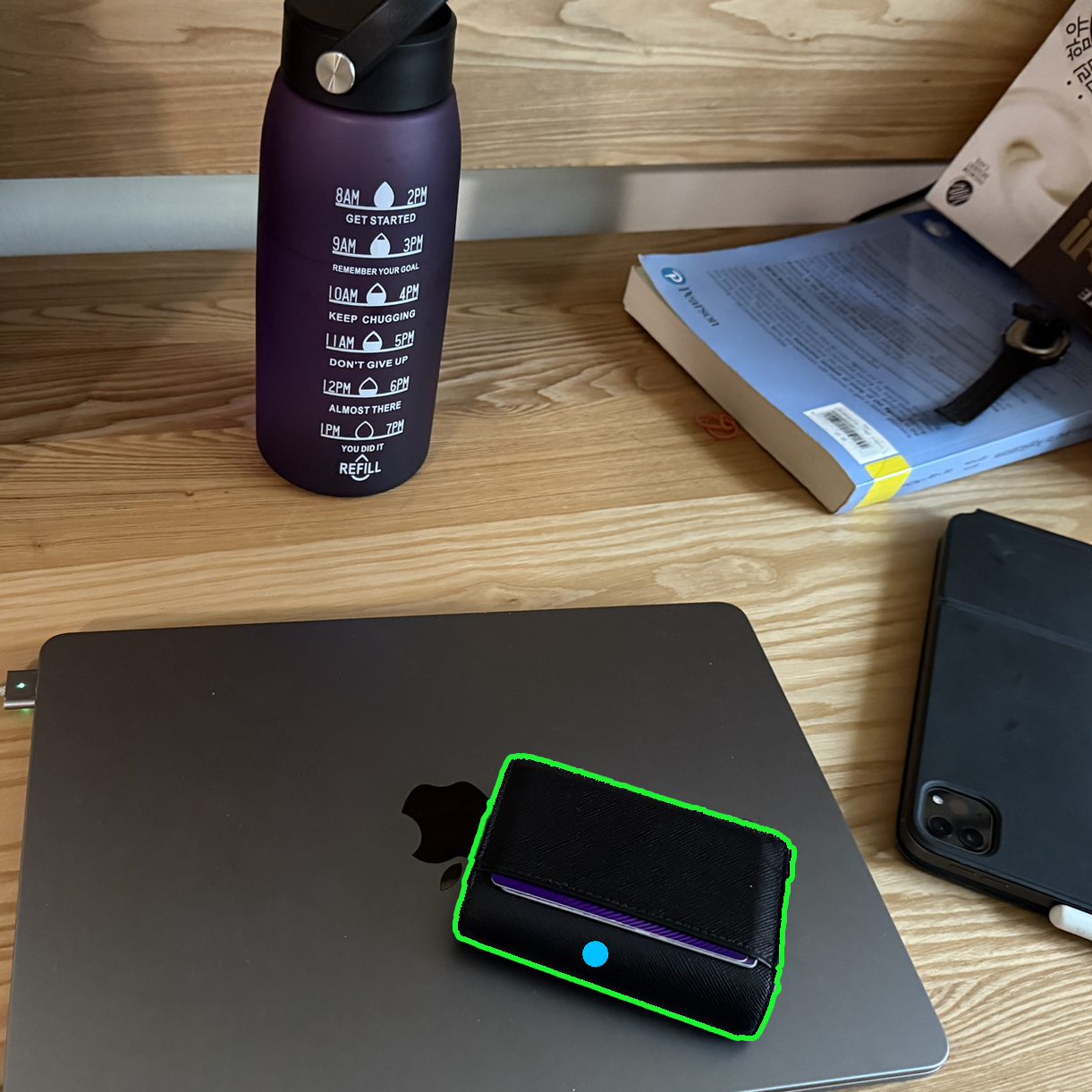}
            \end{minipage}\hspace{0.5em}
            \begin{minipage}{0.3\linewidth}\centering
                \qimg{Which object should you move first to grab the cookie box?}{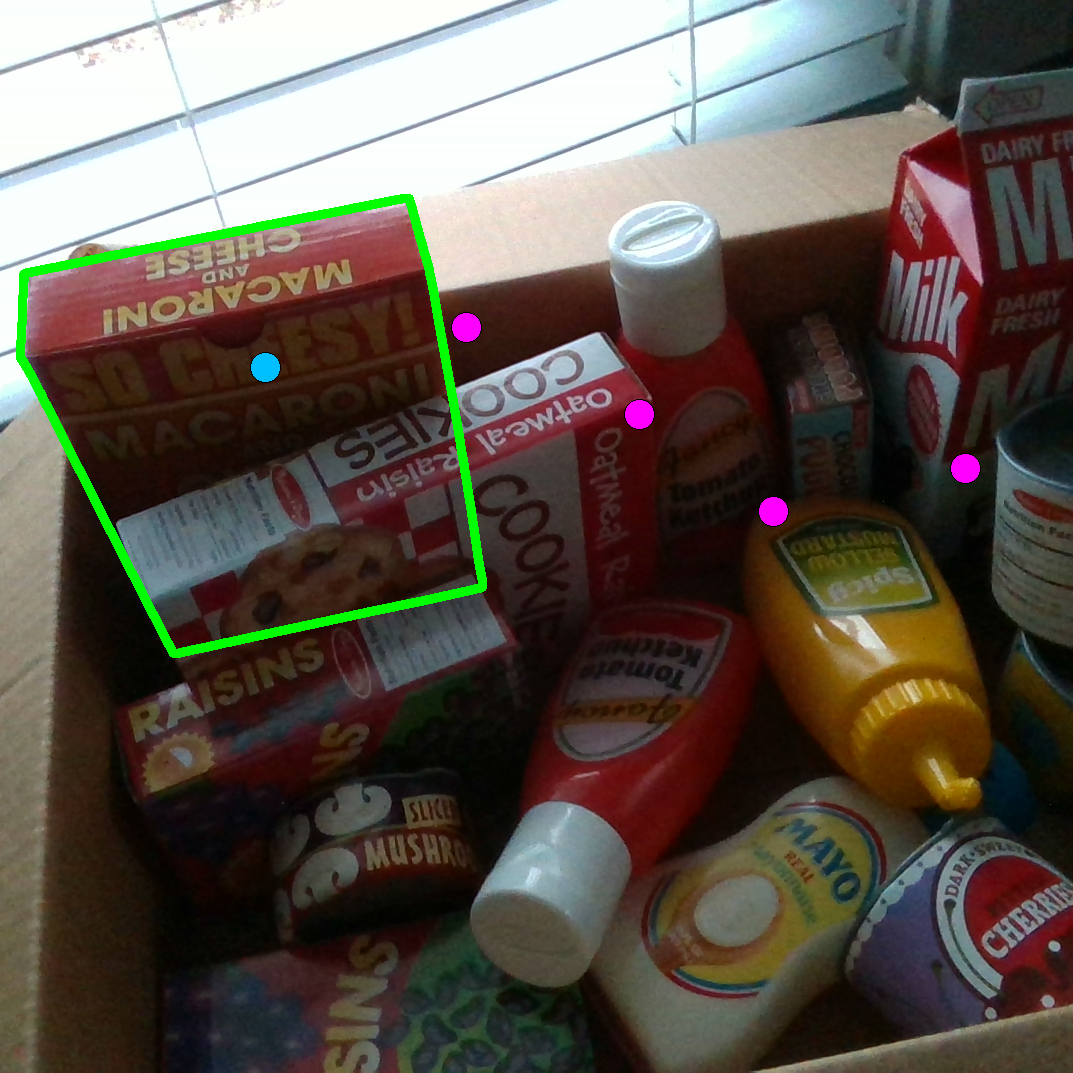}
            \end{minipage}\hspace{0.5em}
            \begin{minipage}{0.3\linewidth}\centering
                \qimg{Which object should you move to access the red bowl?}{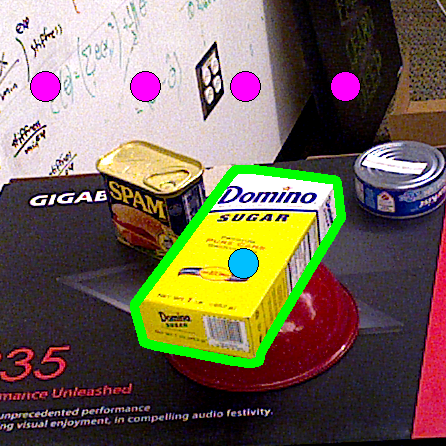}
            \end{minipage}
        \end{minipage} \\\hline\noalign{\vskip 4pt}
        \rotatebox{90}{\textbf{Spatial Reasoning}} &
        \begin{minipage}{\linewidth}\centering
            \begin{minipage}{0.3\linewidth}\centering
                \qimg{Is the drill right of the dry-erase marker?}{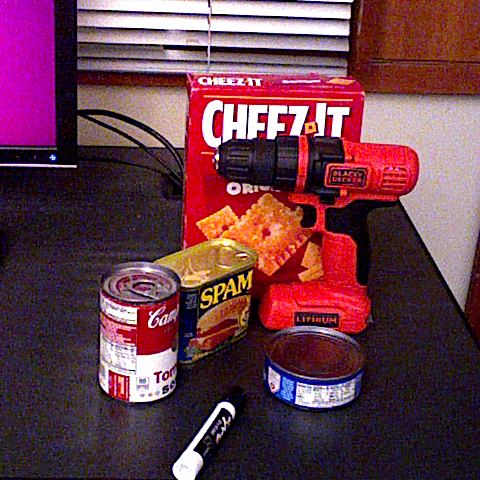}\\[-0.1em]\scriptsize \textcolor{greenline}{\underline{\textcolor{black}{GT: No}}}, \textcolor{magline}{\underline{\textcolor{black}{NVILA: Yes}}}, \textcolor{blueline}{\underline{\textcolor{black}{NVILA SFT: No}}}
            \end{minipage}\hspace{0.5em}
            \begin{minipage}{0.3\linewidth}\centering
                \qimg{Is the red hammer left of the metal skimmer?}{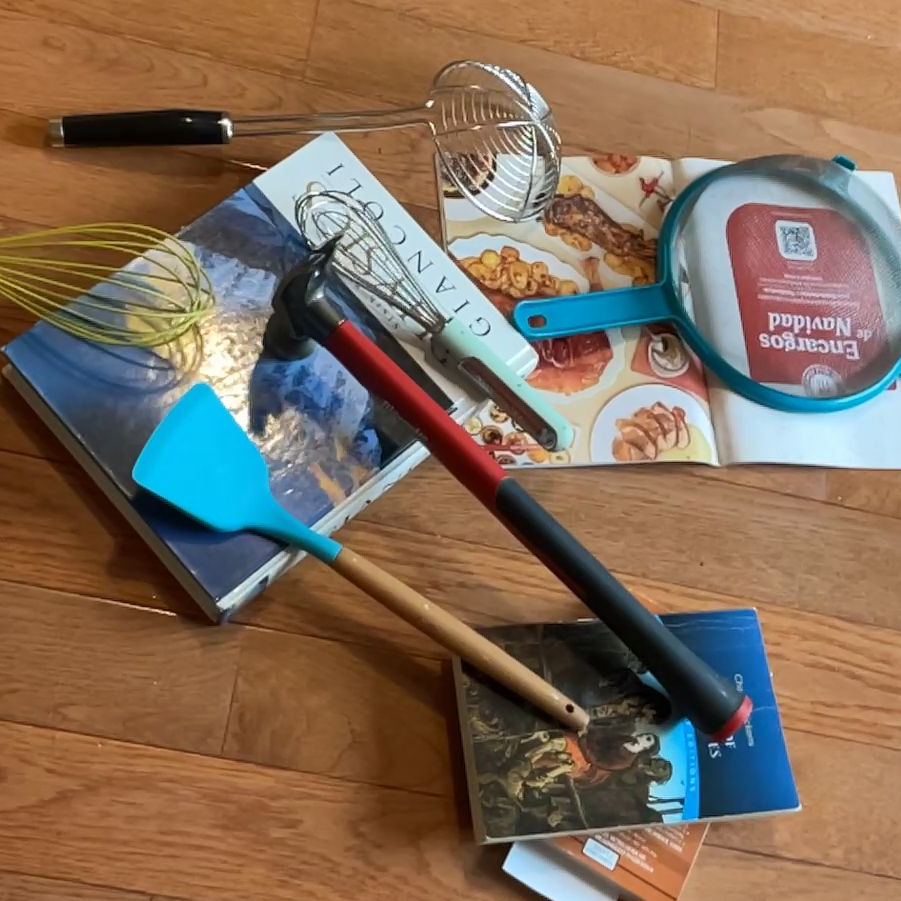}\\[-0.1em]\scriptsize \textcolor{greenline}{\underline{\textcolor{black}{GT: No}}}, \textcolor{magline}{\underline{\textcolor{black}{NVILA: No}}}, \textcolor{blueline}{\underline{\textcolor{black}{NVILA SFT: No}}}
            \end{minipage}\hspace{0.5em}
            \begin{minipage}{0.3\linewidth}\centering
                \qimg{Is the apple further from the camera than the green box? (lab)}{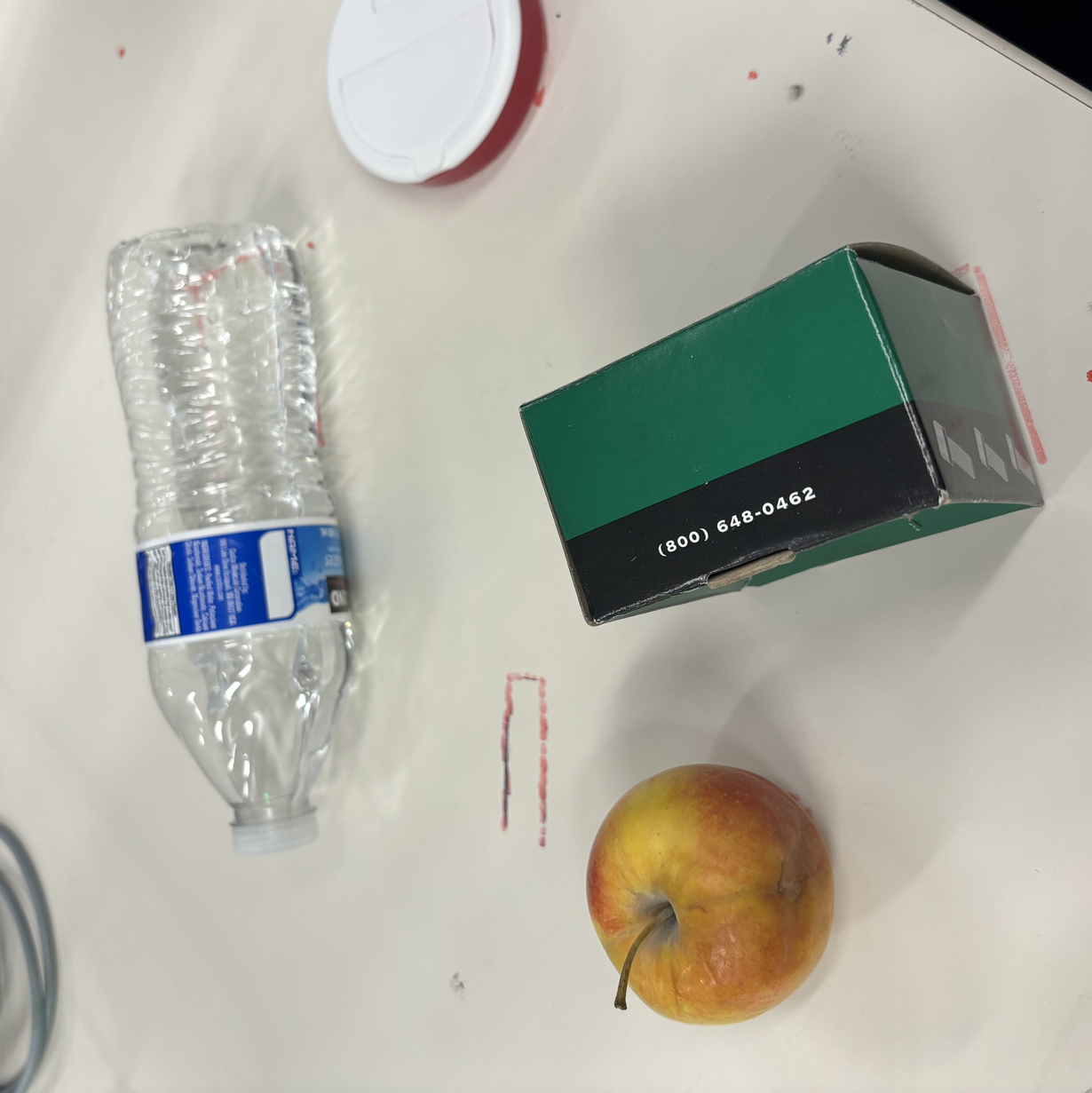}\\[-0.1em]\scriptsize \textcolor{greenline}{\underline{\textcolor{black}{GT: No}}}, \textcolor{magline}{\underline{\textcolor{black}{NVILA: Yes}}}, \textcolor{blueline}{\underline{\textcolor{black}{NVILA SFT: No}}}
            \end{minipage}
        \end{minipage} \\\noalign{\vskip 2pt}\hline
        
    \end{tabular}
    \captionof{figure}{
    Predictions from samples in \datasetname{}-core and \datasetname{}-lab (identified by (lab)), showing improvements gained from fine-tuning on \datasetname.     
    Predictions from \textcolor{magline}{\underline{\textcolor{black}{NVILA (shown in magenta)}}} and \textcolor{blueline}{\underline{\textcolor{black}{NVILA SFT (shown in blue)}}} are shown alongside the \textcolor{greenline}{\underline{\textcolor{black}{Ground Truth (in green)}}}. For the 'Rearrangement' task, the Ground Truth shape delineates the area of valid predictions. Absence of a colored prediction indicates none was made or it was out of frame.
    Images are from HOPE~\cite{tyree2022hope},  HANDAL~\cite{handal}, and YCB-V~\cite{posecnn}.
    }
    \label{fig:qual-comparisons}
    
\end{table*}

\section{Experiments}

We evaluate a diverse set of proprietary and open-source VLMs on our test suites to quantify performance gaps in the novel spatial reasoning and object interaction skills introduced in \datasetname{}. In addition, we fine-tune two open-source VLMs and demonstrate the performance gains and downstream utility of training on our dataset.

\subsection{Setup and Baselines}
We benchmark a range of widely adopted VLMs, including Molmo~\cite{deitke2024molmopixmoopenweights}, Qwen-VL~2.5~\cite{Qwen2.5-VL}, NVILA~\cite{Liu_2025_CVPR}, 
%
%
RoboRefer~\cite{zhou2025roboreferspatialreferringreasoning}, and Intern-VL~\cite{chen2024internvl}. For proprietary systems, we use GPT-5~\cite{GPT5} and Gemini~1.5~ER~\cite{geminiER15} as closed-source baselines.  
We fine-tune Qwen-VL~2.5 and NVILA using their official training codebases on \datasetname{}. Closed-source models are evaluated via their official APIs, while open-source models are tested using their publicly released implementations. All open-source model inferences are run on a single NVIDIA~A100~GPU, and fine-tuning experiments are conducted on a cluster of eight A100~GPUs with default hyperparameters.

\vspace{-5pt}

\subsection{Human Benchmark}  
\vspace{-5pt}
While humans can typically answer simple yes/no questions on spatial or depth reasoning with high accuracy, the images in \datasetname{} present substantial challenges due to heavy clutter, occlusions, and complex scene layouts. Moreover, tasks such as object pose estimation, grasp prediction, and trajectory reasoning require an understanding of \textit{semantic 3D stability}. For example, the appropriate grasp pose for a knife depends on task intent—grasping by the blade for a handover versus by the handle for cutting.  
To establish a comprehensive human performance baseline on \datasetname{}-core, we developed a simple annotation interface for the images in the test set and distributed it to anonymous contributors. Participants were asked to complete tasks such as marking precise 3D object poses, annotating grasp positions, drawing inter-object motion trajectories, identifying which objects must be moved to grasp a target object in a cluttered scene, and answering binary yes/no questions for spatial reasoning.
%
Human annotations were evaluated using the same metrics applied to the VLM baselines, enabling direct comparison across all six object-interaction reasoning tasks. 
1037 responses were gathered from 40+ contributors, covering all 688 VQAs in \datasetname{}-core. 
Each participant annotated between 10 to 25 questions spanning different task types to ensure balanced coverage. 
Roughly 30\% of the questions were answered by multiple annotators, and their scores were averaged to obtain a single human baseline per question. 
We computed mean performance across tasks to derive the human benchmark reported in Table \ref{tab:mainresults}.
\vspace{-5pt}

\subsection{Results}
\vspace{-5pt}
\paragraph{\datasetname{}-core.}  
Training on \datasetname{} leads to consistent performance gains across all six evaluated tasks, as summarized in Table~\ref{tab:mainresults}. 
For \textit{object pose estimation}, open-source VLMs typically default to predicting 2D bounding boxes rather than full 3D cuboid poses. Fine-tuning on our dataset enables NVILA and Qwen-VL to adapt to this richer prediction space while preserving their inherent visual grounding capabilities, resulting in more precise and physically consistent pose predictions.  \textit{Trajectory generation} and \textit{grasp prediction} emerge as challenging skills, primarily because existing pre-training corpora lack supervision for continuous spatial motion or 6-DoF grasp understanding. Training on \datasetname{} exposes these models to explicit geometric cues, improving their ability to reason about motion continuity and feasible grasp locations. While Gemini and Molmo exhibit strong performance on pointing-based tasks, they occasionally misidentify objects in scenes with high clutter or visual distractors.  \textit{Spatial reasoning} and \textit{relative depth perception} tasks show higher accuracies, as these typically reduce to binary or comparative judgments (e.g., “left of,” “closer”), which are well-aligned with the relational reasoning priors acquired during pre-training. Finally, \textit{object rearrangement} remains a difficult task: the best-performing models only reach about 60\% accuracy. This task demands fine-grained understanding of object–object relationships, 3D coordinate alignment, and clutter dynamics—highlighting a key opportunity for future progress in physically grounded reasoning. Figure~\ref{fig:qual-comparisons} presents qualitative examples across all six tasks, comparing predictions from two representative models against the ground truth.  Examples across different rows illustrate that while the baseline NVILA model (magenta) struggles with complex reasoning tasks, fine-tuning on \datasetname{} enables accurate geometric priors (blue) essential for robotic manipulation.
\vspace{-12pt}

\paragraph{\datasetname{}-lab \& out of distribution}

\begin{table}[ht]
\vspace{-10pt}
\centering
\setlength{\tabcolsep}{1pt} 
\renewcommand{\arraystretch}{0.92} 
\small 
\caption{
Comparing models trained on \datasetname{} with baselines on OOD test sets. 
RS-H~\cite{song2025robospatial}, CV-B~\cite{tong2024cambrian1}, SB~\cite{cai2024spatialbot}. 
RoboSpatial results are from original paper. ``--'' denotes unavailable scores. 
Metrics: Pose (IoU), Grasping (Gr; NCE~$\downarrow$), Trajectories (Traj; success rate), Spatial–Depth (S-D; accuracy).
}
\label{tab:ood_testing}
\vspace{-5pt}
\begin{tabular}{>{\raggedright\arraybackslash}lllllllll}
\toprule
\textbf{Models} & 
\textbf{RS-H} & 
\textbf{CV-B} & 
\textbf{SB} &
\multicolumn{4}{c}{\textbf{\datasetname{}-lab}} \\
\cmidrule(lr){5-8}
 & & & & \textbf{Pose} & \textbf{Gr} & \textbf{Traj} & \textbf{S-D} \\
\midrule
RoboSpatial~\cite{song2025robospatial} & 78.0 & - & \underline{64.7} & - & - & - & - \\
NVILA~\cite{Liu_2025_CVPR}              & 63.4 & 78.2 & 47.5 & 6.1 & 4.2 & 0 & 70.0 \\
\quad + \datasetname{}                       & 69.1~\textcolor{green!60!black}{$\uparrow$} & \underline{89.3}~\textcolor{green!60!black}{$\uparrow$} & 50.0~\textcolor{green!60!black}{$\uparrow$} & \underline{16.2}~\textcolor{green!60!black}{$\uparrow$} & \textbf{1.1}~\textcolor{green!60!black}{$\uparrow$} & \underline{28.2}~\textcolor{green!60!black}{$\uparrow$} & \underline{81.2}~\textcolor{green!60!black}{$\uparrow$} \\
Qwen-VL 2.5~\cite{Qwen2.5-VL}         & \underline{78.1} & 88.8 & 60.0 & 12.6 & 3.6 & 0 & 74.4 \\
\quad + \datasetname{}                       & \textbf{81.3}~\textcolor{green!60!black}{$\uparrow$} & \textbf{92.4}~\textcolor{green!60!black}{$\uparrow$} & \textbf{65.0}~\textcolor{green!60!black}{$\uparrow$} & \textbf{25.3}~\textcolor{green!60!black}{$\uparrow$} & \underline{1.3}~\textcolor{green!60!black}{$\uparrow$} & \textbf{37.1}~\textcolor{green!60!black}{$\uparrow$} & \textbf{85.8}~\textcolor{green!60!black}{$\uparrow$} \\
\bottomrule
\end{tabular}
\end{table}

Table~\ref{tab:ood_testing} reports the improvement observed on models fine-tuned on \datasetname{} and tested on \datasetname-lab. 
We observe improvement on all metrics for both Qwen-VL~2.5 and NVILA, demonstrating that training on \datasetname{} enhances 3D grounding and manipulation capabilities, while underscoring remaining challenges of open-world grounded interaction, 
see Figure~\ref{fig:qual-comparisons} for examples taken from \datasetname-lab.
Although \datasetname{} focuses on indoor scenes, its spatial and depth reasoning tasks enable strong domain transfer to out-of-domain benchmarks. 
We evaluate fine-tuned models on three popular spatial reasoning datasets—\textit{Configuration} from RoboSpatial-HOME~\cite{song2025robospatial}, 
\textit{Relation} task from CV-Bench~\cite{tong2024cambrian1}, 
and \textit{Reach} task from SpatialBench~\cite{cai2024spatialbot} 
— chosen for their close alignment with the spatial reasoning tasks in \datasetname{} such as \textit{``Is the picture above the desk?''}.  
Table~\ref{tab:ood_testing} (left-side) reports the results for the RoboSpatial~\cite{song2025robospatial} baseline, and Qwen-VL~2.5~\cite{Qwen2.5-VL} and NVILA~\cite{Liu_2025_CVPR} before and after fine-tuning on \datasetname{}.
Fine-tuned models achieve consistent accuracy gains, highlighting effective zero-shot transfer of spatial reasoning capabilities learnt from \datasetname{}.

\vspace{-15pt}
\paragraph{Robot Experiments}
To evaluate real‐world performance, we deployed models on 15 pick-and-place tasks executed by a Franka robot (e.g., “Move the tomato sauce can to the location of the green box”).
Across all tasks, the base NVILA failed to complete any, whereas NVILA fine-tuned on \datasetname{} succeeded on 10 out of 15. Additional details are provided in the supplementary material.

\begin{table}[ht]
\centering
\vspace{-5pt}
\setlength{\tabcolsep}{0.7pt} 
\renewcommand{\arraystretch}{0.92} 
\small 
\caption{
Ablations on incremental training of NVILA on images within \datasetname{}. 
\textbf{BA-YCBV:} YCB-V only; 
\textbf{BA-YCBV+H:} YCB-V + HANDAL; 
\textbf{BA-YCBV+H+L:} YCB-V + HANDAL + LINEMOD; 
\textbf{BA-NoSpatDep:} \datasetname{} excluding spatial and depth (yes/no) questions. 
}

\label{tab:ablations-data}
\vspace{-5pt}
\begin{tabular}{l
                c
                cc
                c
                c
                c
                c}
\toprule
\multirow{2}{*}{\textbf{Method}} &
\multicolumn{1}{c}{\textbf{Pose}} &
\multicolumn{1}{c}{\textbf{Traj.}} &
\multicolumn{1}{c}{\textbf{Grasps}} &
\multicolumn{1}{c}{\textbf{Spat.}} &
\multicolumn{1}{c}{\textbf{Depth}} &
\multicolumn{1}{c}{\textbf{Rearr.}} \\
 & \textbf{IOU $\uparrow$} & \textbf{SR $\uparrow$} & \textbf{NCE $\downarrow$} & \textbf{SR $\uparrow$} & \textbf{SR $\uparrow$} & \textbf{Rec. (\%) $\uparrow$} \\
\midrule
NVILA (Base)   & 6.5 & 0 & 8.15 & 65.5 & 52.5 & 8.3\\
+ BA-YCBV    & 31.7 & 24.2 & 6.37 & 65 & 75.0 & 16.4\\
+ BA-YCBV+H           & 54.4 & 30.8 & 2.85 & 87.5 & 90.2 & 21.8 \\
+ BA-YCBV+H+L           & 67.2 & \underline{51.8} & 2.02 & \underline{93.5} & \underline{92.6} & 39.2 \\
+ \datasetname{}           & \underline{77.4} & \textbf{64.2} & \textbf{1.40} & \textbf{94.2} & \textbf{94.6} & \textbf{57.7} \\
+ BA-NoSpatDep & \textbf{78.2} & 50.0 & \underline{1.69} & 62.5 & 50.6 & \underline{50.3} \\
\bottomrule
\end{tabular}
\vspace{-20pt}
\end{table}

\paragraph{Ablations on data recipe.}  
We evaluate the effect of combining data from multiple BOP families—HOPE~\cite{tyree2022hope}, HANDAL~\cite{handal}, YCB-V~\cite{posecnn}, and LINEMOD~\cite{Hinterstoier2012ModelBT}—and analyze performance across question types (Table~\ref{tab:ablations-data}). Performance consistently improves as additional datasets are introduced, suggesting that increased visual and semantic diversity—spanning object geometries, textures, and spatial layouts—enhances generalization in fine-grained spatial reasoning.  
The base model initially fails in trajectory questions but quickly adapts during fine-tuning, reaching 24\% accuracy using only VQAs from YCB-V. This underscores spatial pre-training, which provides transferable visual grounding for new tasks. Subsequent inclusion of additional datasets further boosts performance across all skills. Notably, removing binary spatial and depth reasoning tasks (\textit{BA-NoSpatDep}) leads to a drop in overall performance, confirming that such auxiliary tasks strengthen object-interaction reasoning through multi-skill co-training.

\vspace{-5pt}

\section{Conclusion}
\vspace{-5pt}

We presented \datasetname{}, a large-scale training dataset that introduces object-interaction reasoning. It includes tasks such as 3D object poses, grasp affordances, motion trajectories, and object rearrangements. 
%
In addition, we introduced two complementary benchmarks, \datasetname{}-core and \datasetname{}-lab, 
to evaluate models in seen and unseen environments.  
Experiments on in-domain and out-of-domain test sets show the rich spatio-geometric priors encoded in \datasetname{} enhance trained models ability to reason and act upon objects in complex scenes. 

\section{Acknowledgments}
We are grateful to Huy Le for supporting an earlier stage of this dataset. 
We also thank Naren Devarakonda, Amrita Mazumdar, Alex Zook, Siyi Chen, Chan Hee (Luke) Song, and Faisal Ladhak for their helpful input on this manuscript. Also thank you to all the anonymous participant to our human study. This paper is supported in part 
by the New York University Abu Dhabi (NYUAD) Center for Artificial Intelligence and Robotics,
funded by Tamkeen under the NYUAD Research Institute Award CG010, and by NSF under grant
number 2208189.
{
    \small
    \bibliographystyle{ieeenat_fullname}
    \bibliography{main}
}



\end{document}